\documentclass{article}

\usepackage{microtype}
\usepackage{graphicx}
\usepackage{subcaption}
\usepackage{booktabs} 
\usepackage{tabularray}
\UseTblrLibrary{booktabs}

\usepackage{hyperref}

\usepackage[preprint]{icml2026}

\usepackage{amsmath}
\usepackage{amssymb}
\usepackage{mathtools}
\usepackage{amsthm}

\usepackage[capitalize,noabbrev]{cleveref}
\usepackage{enumitem}
\usepackage{multirow}
\usepackage{graphicx}
\usepackage{wrapfig}
\usepackage{array}
\usepackage{pifont}
\usepackage{bbm}
\usepackage[table,xcdraw]{xcolor}
\usepackage{titletoc} 
\usepackage{algorithm}
\usepackage{algorithmic}
\usepackage{xspace}

\theoremstyle{plain}

\theoremstyle{definition}

\theoremstyle{remark}

\usepackage[textsize=tiny]{todonotes}

\icmltitlerunning{Your CLIP has 164 dimensions of noise}

\begin{document}

\twocolumn[
  \icmltitle{Your CLIP has 164 dimensions of noise: Exploring the embeddings covariance eigenspectrum of contrastively pretrained vision-language transformers.}



  \icmlsetsymbol{equal}{*}

  \begin{icmlauthorlist}
    \icmlauthor{Jakub Grzywaczewski}{equal,wut,ccai}
    \icmlauthor{Dawid Płudowski}{equal,wut,ccai}
    \icmlauthor{Przemysław Biecek}{wut,ccai,uw}
  \end{icmlauthorlist}

  \icmlaffiliation{wut}{Warsaw University of Technology}
  \icmlaffiliation{ccai}{Centre for Credible Artificial Intelligence}
  \icmlaffiliation{uw}{University of Warsaw}

  \icmlcorrespondingauthor{Jakub Grzywaczewski}{jakubzgrzywaczewski@gmail.com}

  \icmlkeywords{Machine Learning, ICML}

  \vskip 0.3in
]



\printAffiliationsAndNotice{\icmlEqualContribution}

\begin{abstract}
  Contrastively pre-trained Vision-Language Models (VLMs) serve as powerful feature extractors.
  Yet, their shared latent spaces are prone to structural anomalies and act as repositories for non-semantic, multi-modal noise.
  To address this phenomenon, we employ spectral decomposition of covariance matrices to decompose the VLM latent space into a multi-modal semantic signal component and a shared noise subspace.
  We observe that this noise geometry exhibits strong subgroup invariance across distinct data subsets. 
  Crucially, pruning these shared noise dimensions is mainly harmless, preserving or actively improving downstream task performance.
  By isolating true semantic signals from artifactual noise, this work provides new mechanistic insights into the representational structure of modern VLMs, suggesting that a substantial fraction of their latent geometry is governed by shared, architecture-level noise rather than task-relevant semantics alone. 
\end{abstract}

\section{Introduction}

Contrastively pre-trained Vision-Language Models (VLMs) \citep{oord2018representation, chen2020simple, radford2021learning, zhai2023sigmoid} are deployed globally as feature extractors across a wide variety of tasks from zero-shot classification \citep{radford2021learning} and cross-modal retrieval \citep{kordopatis2025ilias} to text-based segmentation \citep{yu2023convolutions} and visual question answering \citep{song2022clip}.
However, the widespread reliance by both researchers and industry on the rich context provided by these shared latent spaces exposes an intriguing point of failure: these representations often begin to encode non-semantic, uninformative noise originating from both modalities.

\begin{figure}[t]
    \centering
    \includegraphics[width=1\linewidth]{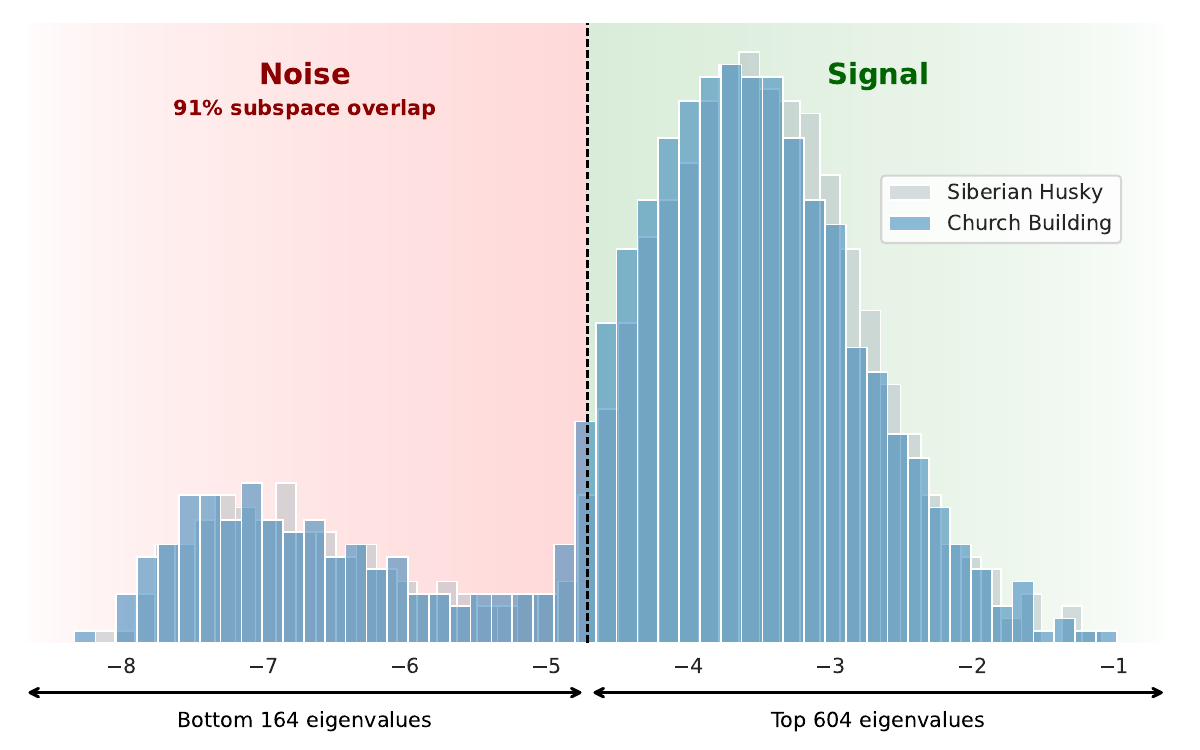}
    \caption{The eigenspectra of sample covariance matrices, computed from CLIP ViT-L/14 embeddings for two unrelated ImageNet classes, are highly similar (shown here as the $\log_{10}$ of the eigenvalues). Despite no apparent correlation nor common samples, their identified class-specific noise directions exhibit strong overlap, achieving mSCSA of $91\%$ (Equation~\ref{p:mscsa}).
    }
    \label{fig:figure1}
\end{figure}

To address this vulnerability, extensive research has sought to map the geometry of these latent spaces \citep{mikolov2013distributed, arditi2024refusal, lee2025shared}, uncovering structural phenomena such as the linear representation hypothesis \citep{park2024linear}, the hierarchical nature of concepts \citep{park2025geometry}, and the modality gap \citep{liang2022mind, chowers2026modality, fahims2025itsnot, yu2026modality, mistretta2025cross}.
The urgency of this geometrical analysis is underscored by a broader push for the mechanistic decomposition of neural networks \cite{sharkey2024openproblems}, in this case isolating true semantic signals from artifactual noise.
A~critical prior work context for this challenge is dimensional collapse \citep{jingunderstanding}, a phenomenon initially observed in single-modal contrastively-learned models like ResNet-50 \citep{he2016deep}.
While contrastive learning prevents the entire latent space from collapsing, it still allows individual dimensions to collapse. 

In this short paper, we demonstrate that dimensional collapse persists in modern VLMs such as CLIP and SigLIP, albeit in an altered form. 
Rather than collapsing entirely, these dimensions become repositories for the model to encode irrelevant data.
Following~\citet{jingunderstanding}, we analyze these dimensions through the spectral decomposition of covariance matrices computed over large-scale datasets \citep{deng2009imagenet, schuhmann2022laion} of different~modalities.

\textbf{The primary contribution lies in formally decomposing the VLM latent space into the shared noise subspace and multi-modal signal}.
Specifically, we observe a striking alignment of subspaces within the noise component, which constitutes the lower bulk of the covariance eigenvalues \citep{marvcenko1967distribution}.
As another contribution, we demonstrate that pruning these shared noise dimensions is mainly harmless; it does not degrade downstream performance and, in some instances, improves it. 
Finally, we show that this geometry exhibits strong subgroup invariance. We observe that the eigenspectra of covariance matrices computed over distinct ImageNet \citep{deng2009imagenet} classes remain highly similar, as highlighted in Figure~\ref{fig:figure1}, and that the eigenvectors corresponding to the smallest eigenvalues consistently align with our identified noise dimensions, regardless of the chosen data subset.

\section{Related Work}

The latent space of transformer-based \citep{vaswani2017attention} VLMs, such as the contrastively trained CLIP \citep{radford2021learning} and SigLIP \citep{zhai2023sigmoid, tschannen2025siglip} models, encodes concepts linearly \citep{park2024linear} and hierarchically \citep{park2025geometry}, yet remains subject to structural anomalies \citep{mikolov2013distributed, arditi2024refusal, lee2025shared}.
For instance, representations often suffer from the cone effect, where projected data is confined to a narrow cone of the embedding space, as initially observed in BERT \citep{ethayarajh2019contextual}, as well as the modality gap \citep{liang2022mind}, wherein image and text embeddings remain strictly separated throughout the entire process of pre-training.
Subsequent studies \citep{chowers2026modality, fahims2025itsnot, yu2026modality} have linked this modality gap directly to the foundational contrastive alignment objective \citep{oord2018representation, chen2020simple} and to underlying dimensional collapse \citep{jingunderstanding}.

To refine these geometries, recent literature has explored targeted subspace interventions.
For example, \citet{zhuenhancing} isolate the subspace spanned by a subset of class-name text representations to prune uninformative features.
Similarly, IsoCLIP \citep{magistri2026isoclip} filters CLIP dimensions by isolating the isotropic region of the intra-modal singular values, while \citet{ospanov2025scendi} utilize the Schur complement of the empirical covariance matrix to enhance the diversity of prompt-guided generation. 

Furthermore, \citet{betserwhitened} suggest using CLIP as a surrogate for likelihood via whitening (computing the inverse square root of the covariance matrix).
This establishes a Mahalanobis distance metric where samples projecting onto the smaller eigenvalues exhibit a higher norm, thereby corresponding to lower likelihood.
Our work mechanistically extends this concept by demonstrating that these lower eigenvalues explicitly encode the shared noise dimensions.

\begin{table*}[t]
\centering
\caption{Effects of projecting the embeddings of different vision backbones on downstream tasks: ImageNet zero-shot image classification and LAION-2B image–text alignment, where classification performance is measured by Top-5 accuracy and alignment is quantified by the change in average cosine similarity.}
\label{tab:top5_acc}
\renewcommand{\arraystretch}{1.15}
\setlength{\tabcolsep}{10pt}

\begin{tblr}{
    colspec = {lcccccc},
    vline{2,4,7} = {dashed},
    width = \textwidth,
    row{1,2,3} = {font=\bfseries},
    row{2} = {abovesep=4pt, belowsep=2pt},
    row{3} = {abovesep=2pt, belowsep=4pt},
    column{5} = {rightsep=15pt},
    cell{1}{2} = {c=2}{halign=c},
    cell{1}{4} = {c=3}{halign=c},
    cell{2}{6} = {halign=c},
    cell{2}{7} = {halign=c},
    cell{3}{6} = {halign=c},
    cell{3}{7} = {halign=c},
}
\toprule
\textbf{} &
\textbf{Dimension details} &
\textbf{} &
\textbf{ImageNet Zero-shot (Top5 Accuracy)} &
\textbf{} &
\textbf{} &
\textbf{LAION-2B (CosSim)} \\
\midrule
\textbf{Backbone} &
\textbf{Latent size} &
\textbf{Noise size} &
\textbf{Original} &
\textbf{Noise-Free} &
\textbf{Random} &
\textbf{Noise-Free} \\
& \textbf{} & \textbf{} &
\textbf{dimensions} &
dimensions &
\textbf{dimensions} &
\textbf{dimensions} ($\times 10^{2}$) \\
\midrule
CLIP ViT-B/16    & 512  & 21 (4\%)   & $88.5$ & $88.5$ & $87.9_{\pm 0.09}$ & $+0.090_{\pm 0.070}$ \\
CLIP ViT-B/32    & 512  & 14 (3\%)   & $85.2$ & $85.2$ & $84.9_{\pm 0.04}$ & $+0.030_{\pm 0.040}$ \\
CLIP ViT-L/14    & 768  & 164 (21\%) & $91.2$ & $91.1$ & $89.3_{\pm 0.07}$ & $+0.090_{\pm 0.070}$ \\
SigLIP ViT-B/16  & 768  & 118 (15\%) & $92.7$ & $92.6$ & $91.9_{\pm 0.05}$ & $+1.600_{\pm 0.310}$ \\
SigLIP2 ViT-B/16 & 768  & 185 (24\%) & $88.0$ & $88.0$ & $85.2_{\pm 0.22}$ & $+1.590_{\pm 0.380}$ \\
SigLIP2 ViT-L/16 & 1024 & 405 (40\%) & $89.0$ & $88.4$ & $83.9_{\pm 0.56}$ & $+1.000_{\pm 0.280}$ \\
\bottomrule
\end{tblr}
\end{table*}

\begin{figure}[t]
    \centering
    \includegraphics[width=1\linewidth]{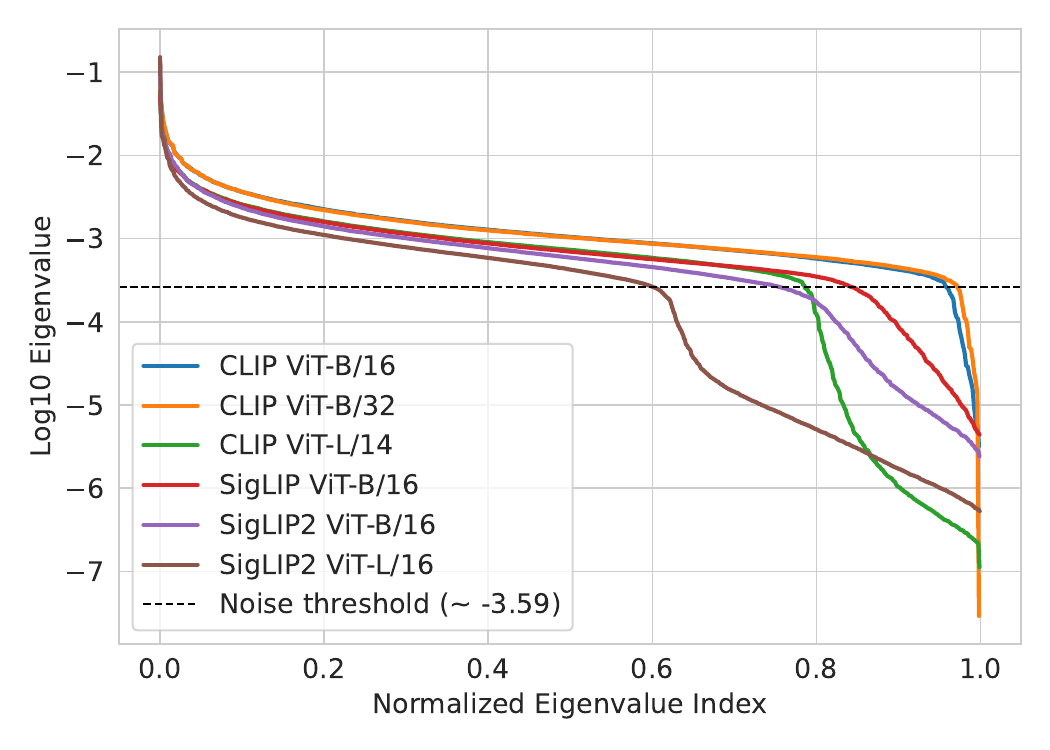}
    \caption{Progression of the $\log_{10}$ of eigenvalues of the average covariance matrix (\Cref{eqn:avg_cov}) for all considered VLM backbones. All evaluated models exhibit a sharp dip in eigenvalues immediately after the shared noise threshold.}
    \label{fig:figure2}
    \vspace{-1em}
\end{figure}

\section{Methodology}
Following \citet{jingunderstanding}, we begin by examining the empirical covariance matrices of the model embeddings. Let $z_i \in \mathbb{R}^d$ represent the model's embedding of an input $x_i$ (either an image or text tokens). The empirical covariance matrix $\Sigma$ is computed over a dataset of $n$ samples as:
\begin{equation}
    \Sigma = \frac{1}{n-1} \sum^{n}_{i=1} (z_i - \bar{z}) (z_i - \bar{z})^T,
\end{equation}
where $\bar{z}$ denotes the empirical mean of the embeddings. We compute these covariance matrices separately for each modality to obtain $\Sigma_{\text{text}}$ and $\Sigma_{\text{img}}$. Unless stated otherwise, all covariance matrices are scaled to have a trace of $1.0$ to allow for direct comparison.

The principal directions of the embedding space are then obtained via spectral decomposition. For instance, the image covariance matrix is decomposed as $\Sigma_{\text{img}} = U_{\text{img}} \Lambda_{\text{img}} U_{\text{img}}^T$, where $U_{\text{img}}$ is an orthonormal matrix whose columns correspond to the eigenvectors of $\Sigma_{\text{img}}$. To analyze the joint latent space, we further define the average cross-modal covariance matrix as
\begin{equation}
\label{eqn:avg_cov}
    \Sigma_{\text{avg}} := \frac{1}{2} \left( \Sigma_{\text{img}} + \Sigma_{\text{text}} \right).
\end{equation}

\paragraph{Isolating Shared Noise Dimensions}
Analyzing the eigenvalues of the image, text, and average covariance matrices across various models reveals a consistent, sharp drop in all curves beyond a specific noise threshold (approximately~$10^{-3.6}$). To prevent biasing our isolation process toward any single modality, we estimate this cutoff threshold on the average covariance matrix (\Cref{fig:figure2}) using the elbow method. The final threshold is the minimum across all models; individual eigenvalue progressions can be found in \Cref{apx:noise_threshold}.

\paragraph{Mean Squared Cosine of Subspace Angles (mSCSA)}
\label{p:mscsa}
To quantitatively measure the overlap between two distinct vector subspaces, we employ the mean squared cosine of their principal (subspace) angles. This metric is derived from the singular value decomposition (SVD) of the overlap matrix $M = V_1^T V_2$, where $V_1$ and $V_2$ are matrices whose $p$ orthonormal columns span the two respective subspaces. The singular values of $M$ correspond exactly to the cosines of the principal angles: $1 \geq \cos(\theta_1) \geq \cos(\theta_2) \geq \cdots \geq \cos(\theta_p) \geq 0$. We define the mSCSA metric as:
\begin{equation}
    \label{eqn:mscsa}
    \text{mSCSA} := \frac{1}{p} \sum^{p}_{i=1} \cos^2(\theta_i).
\end{equation}
This formulation provides a stable, bounded estimation of subspace alignment, yielding $1.0$ for perfectly identical subspaces and $0.0$ for strictly orthogonal ones. Crucially, this metric provides a robust way to assess overlap in high-dimensional, noisy spaces, where simply computing the rank of the concatenated matrix $[V_1, V_2]$ almost always yields a full rank of $2p$ rather than a meaningful measure of intersection.

\section{Experiments and Results}

Our primary hypothesis posits that the latent space of contrastive models like CLIP is partitioned into two distinct components: \emph{task-relevant} signal dimensions and \emph{shared}, \emph{non-semantic} noise dimensions. To rigorously test this, we structure our evaluation around three key objectives: verifying the global nature of these dimensions, assessing their impact on downstream tasks, and qualitatively inspecting their semantic contents.

\paragraph{Experimental Setup}
To estimate the covariance matrices for the text modality, we sample 1 million text descriptions from the \textit{laion2B-en-aesthetic} subset of LAION-2B~\citep{schuhmann2022laion}. 
For the image modality, we utilize the complete ImageNet-1K training set (1.2M images) alongside approximately 250K successfully downloaded images from the aforementioned LAION subset, as downloading the LAION URLs manually often resulted in network failures.

\begin{figure}[h]
    \centering
    \includegraphics[width=0.9\linewidth]{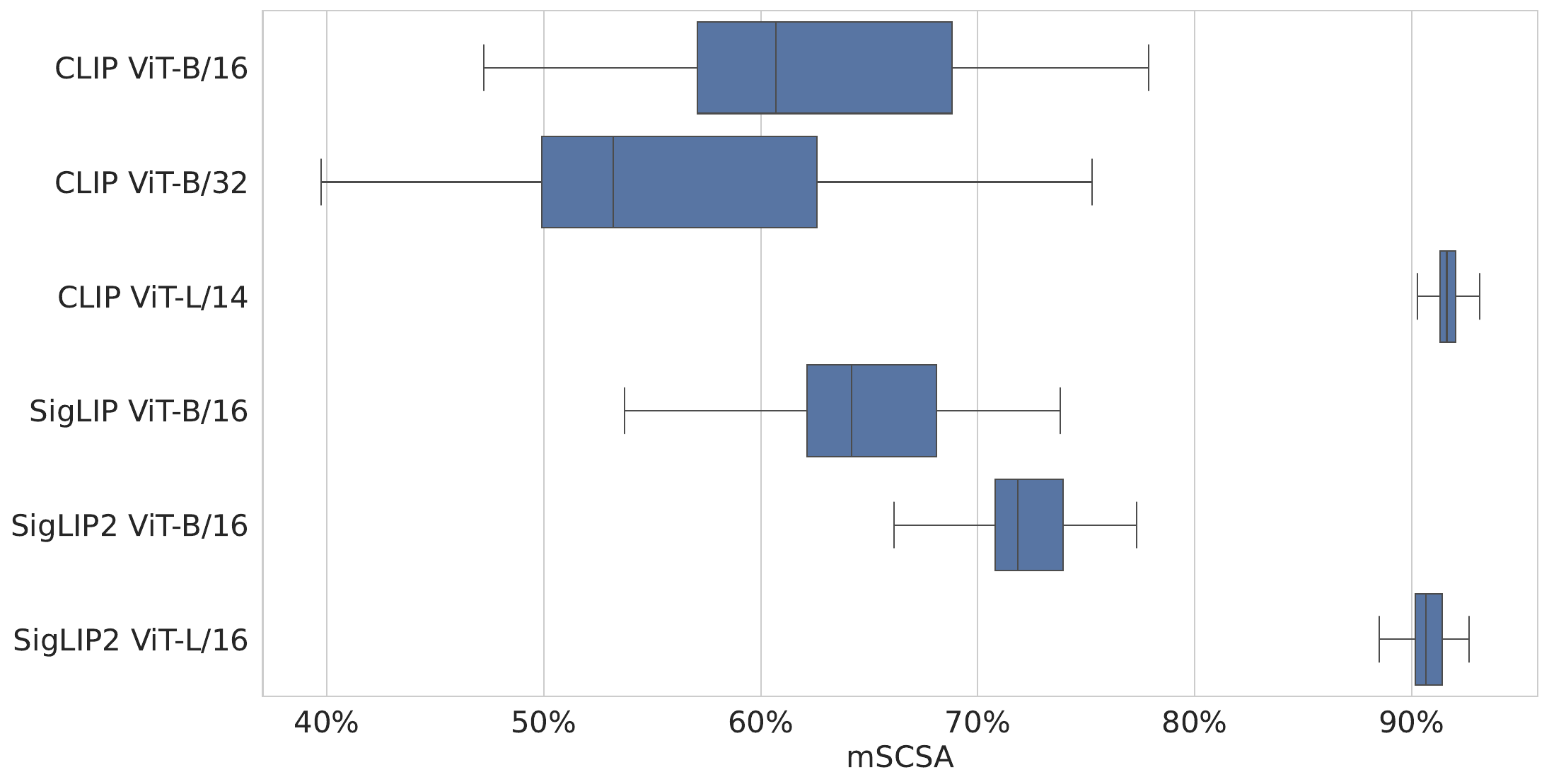}
    \caption{Overlap of the noise dimensions, computed using the average covariance matrix, and the lower eigenvectors of each ImageNet class covariance. For CLIP ViT-L/14 and SigLIP2 ViT-L/16, the biggest models in their families, this overlap is very high at above $90\%$.}
    \label{fig:overlap}
\end{figure}

\paragraph{Universality of Noise Dimensions}
We first seek to confirm that the identified noise vectors are ubiquitous across the dataset rather than class-specific artifacts. To do this, we isolate the $p$ lowest eigenvalues of the covariance matrix computed for \textit{each individual class} in ImageNet. We then compute the Mean Squared Cosine of Subspace Angles (mSCSA) between our globally extracted noise vectors and the eigenvectors of these class-specific low-variance dimensions, where $p$ matches the estimated number of noise directions for the given backbone. As shown in \cref{fig:overlap}, the mSCSA scores (above $50\%$ for most models, and above $90\%$ for the two biggest ones) indicate high alignment of the noise subspaces, which is further strengthened by the fact that an average root mean square distance between mean-centered eigenvalues of different classes is between $0.09$ and $0.19$.

\paragraph{Impact on Downstream Task Utility}Assuming that the found dimensions strictly encode non-semantic noise (e.g., memorized or out-of-distribution artifacts), truncating them should not degrade downstream utility. To verify this hypothesis, we evaluate model performance before and after projecting representations away from this noise subspace. We define a noise-removing projection matrix $P = I - VV^T$, where $V \in \mathbb{R}^{d \times p}$ is the orthonormal basis of eigenvectors corresponding to the $p$ eigenvalues of $\Sigma_{\text{avg}}$ falling below the noise threshold. We evaluate this across two standard tasks: zero-shot image classification on ImageNet, and text-image alignment (measured via cosine similarity) on LAION-2B pairs.

\begin{figure}[t]
    \centering
    \includegraphics[width=0.9\linewidth]{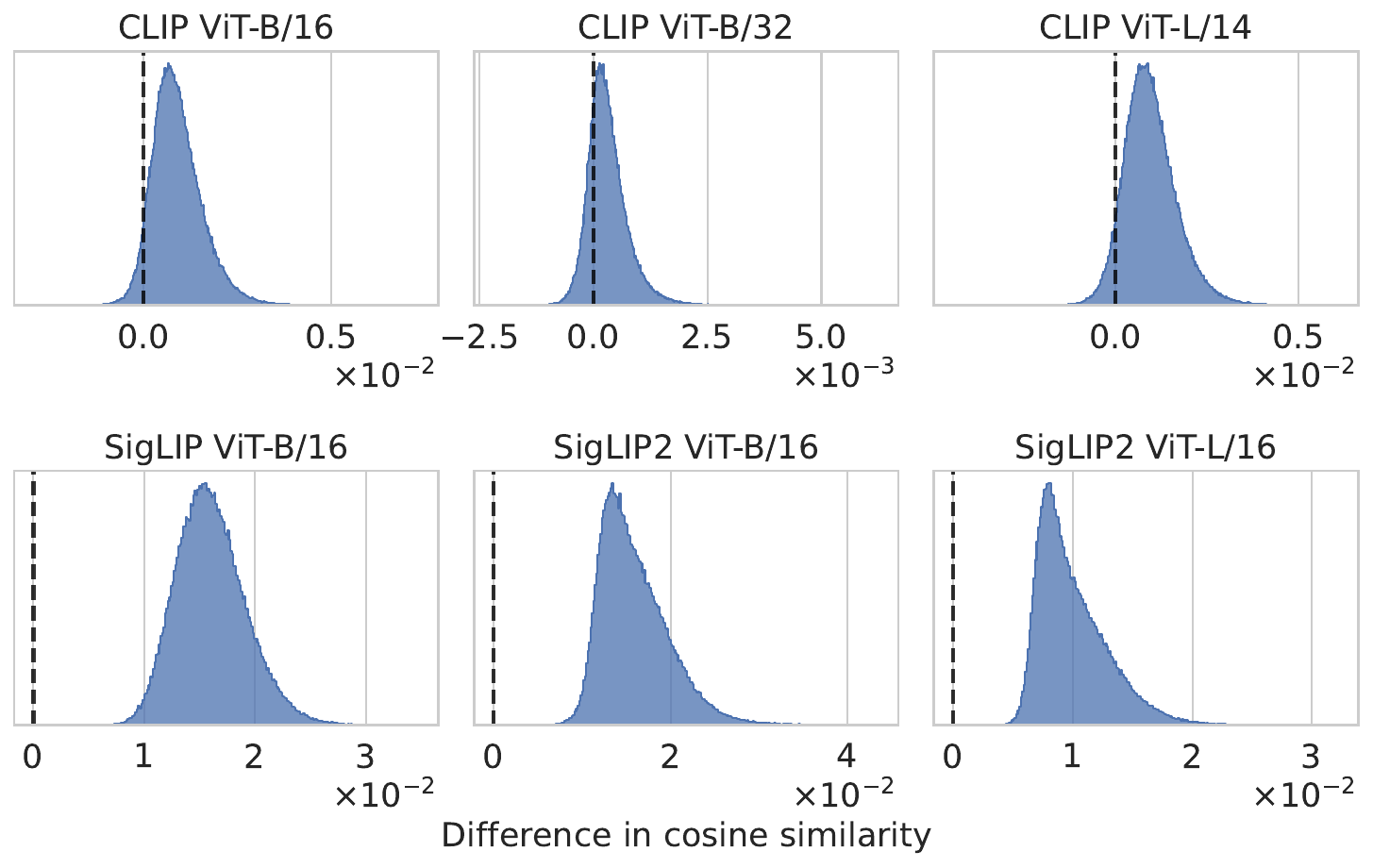}
    \caption{The distribution of the difference between the cosine similarities of image-caption pairs from LAION-2B after and before the noise-removing projection. We can observe that in all cases the significantly largest portion of the probability mass is located on the right of $0$, indicating an increase in similarity.}
    \label{fig:cossim-diff}
\end{figure}

\Cref{tab:top5_acc} demonstrates that removing the noise dimensions results in a negligible performance drop on the task of ImageNet zero-shot classification. We confirm that by running an ablation (also present in the table) in which we randomly remove $p$ directions ($500$ times) from the original basis and check that, compared to the random approach, the noise direction conveys nearly no information. For histograms resulting from this ablation, we refer the reader to \Cref{apx:random_direction}. For the case of text-to-image alignment, we discover a systematic increase in the average cosine similarity (\Cref{tab:top5_acc}) and beyond (\Cref{fig:cossim-diff}), an effect that is amplified as we scale the size of the latent dimension of the model.

\paragraph{Qualitative Inspection of the Noise Subspace}Finally, we conduct a small qualitative analysis to visually and semantically inspect what these shared noise dimensions actually encode. Using CLIP ViT-L/14, selected due to its standard prevalence in zero-shot classification literature \citep{clip_vit_large_patch14}, we project normalized image embeddings onto the noise subspace and compute their norms. We surface the specific samples that exhibit the strongest activations within these dimensions. The results are provided within~\Cref{apx:examples}. As illustrated in~\Cref{fig:noimage}, most verified examples contain only a placeholder for unavailable images.

Furthermore,~\Cref{fig:random} presents some of the few noise-activated images that contain actual content. Interestingly, a strong theme shift can be observed when $\Sigma_{\text{img}}$ is estimated using only ImageNet, excluding LAION images. As indicated in~\Cref{fig:marvel}, images activated in the noise subspace contain Marvel movie topics. While the origin of this behavior is not clear, we argue that it might be an implication of the fact that ImageNet was published before these movies, so they might be seen as out-of-distribution for the estimation. This highlights the importance of using versatile data for covariance estimation.

\newpage
\section{Discussion}
While a uniform noise threshold is conceptually elegant, we must acknowledge that its optimal value likely depends on both the model architecture and the specific data distribution; thus, a per-model or per-modality selection of the cut-off may yield more refined results. The existence of noise subspaces containing uninformative (or memorized) features could be rebutted by noting that the drop in the eigenspectrum is not as instantaneous as the sharp phase transition observed by \citet{jingunderstanding}. However, we argue that the dip exists for a reason beyond numerical instabilities, as different models (even of the same architecture) can exhibit exponential decay (linear in log-space) across almost the entire eigenspectrum. This phenomenon can be for example observed in the image-only model DINOv2 \citep{oquabdinov2} in \Cref{apx:noise_threshold}.

\paragraph{Limitations and Future work}
While our current empirical scope is deliberately focused, we prioritize the analysis of CLIP ViT-L/14, the prevailing standard for zero-shot classification, to ensure our foundational findings are grounded in the community's most widely utilized architecture. Furthermore, preliminary observations of the SigLIP model family reveal intriguing structural properties; specifically, the cardinality of the noise subspace does not scale linearly with the overall embedding dimensionality, suggesting a complex underlying mechanism for how these models allocate latent capacity. Future work will expand this evaluation to a broader spectrum of vision-language backbones to verify the universality of these phenomena. Ultimately, establishing a formal theoretical framework for this mechanism, particularly its potential causal relationship with the emergence of the modality gap in contrastive latent spaces, remains a critical direction for future research.

\newpage
\bibliography{references}
\bibliographystyle{icml2026}

\newpage
\appendix
\crefalias{section}{appendix}

\onecolumn

\section{Noise Threshold Selection}
\label{apx:noise_threshold}

\begin{figure}[h]
    \centering
    \includegraphics[width=1\linewidth]{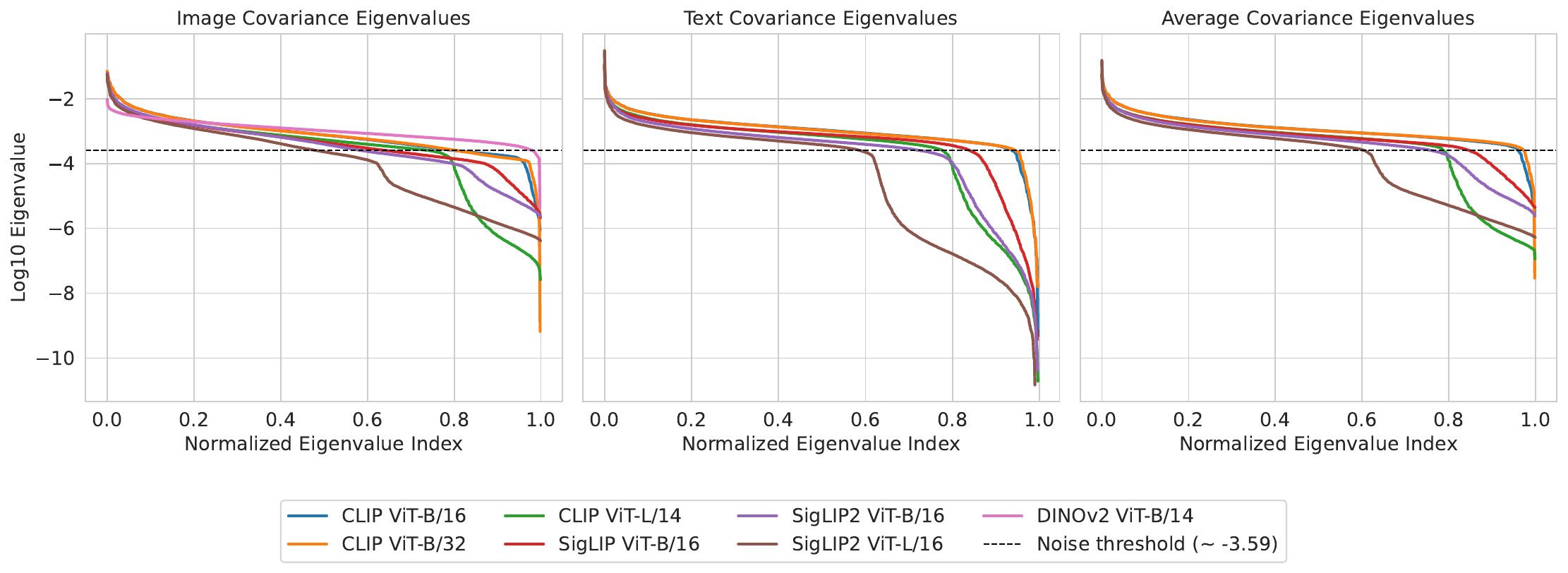}
    \caption{Progression of log eigenvalues for image, text, and average sample covariance matrix with the indicated noise threshold computed as the minimum of knee points for each model.}
    \label{fig:cov_threshold}
\end{figure}

\begin{figure}[h]
    \centering
    \includegraphics[width=1\linewidth]{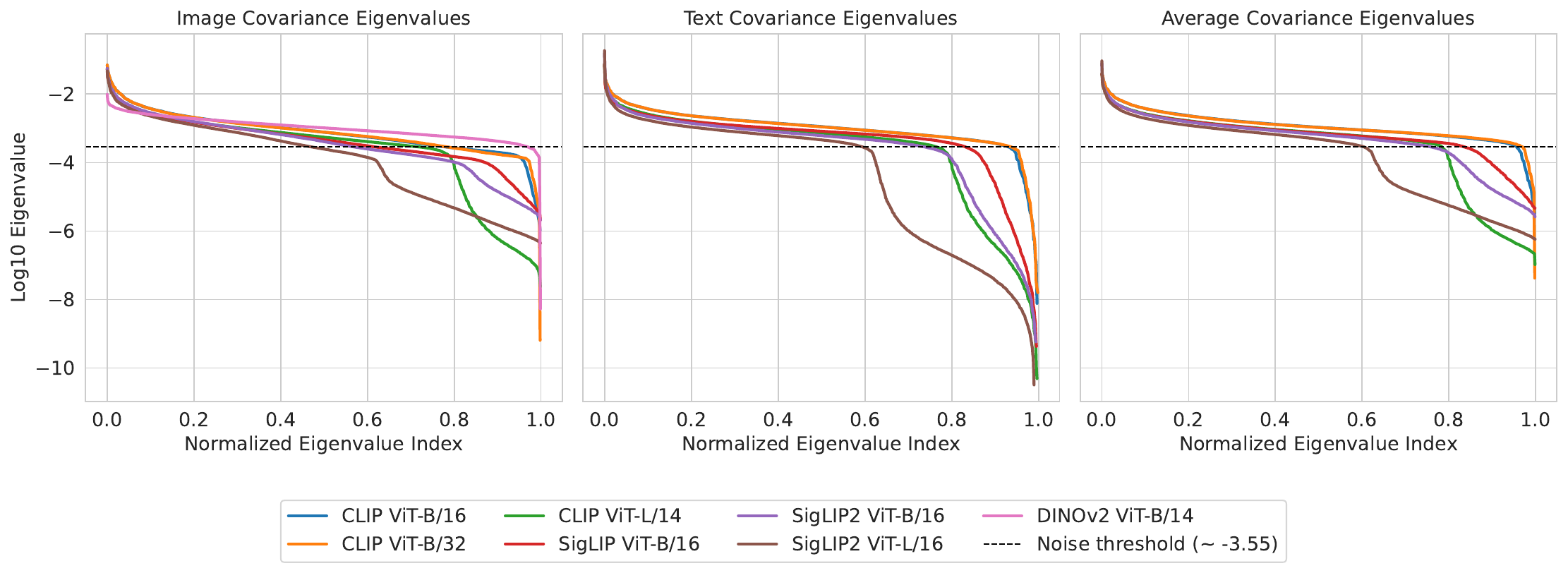}
    \caption{Progression of log eigenvalues for image, text, and average kernel covariance matrix with the indicated noise threshold computed as the minimum of knee points for each model.}
    \label{fig:cosim_cov_threshold}
\end{figure}

In the end we went with the eigenvalues of the sample covariance matrix as a way to establish the noise threshold. Alternatively, as in \cite{ospanov2025scendi} we could have used the eigenspectrum of the kernel covariance matrix defined by the cosine-similarity kernel:
\begin{equation*}
\Sigma = \frac{1}{n-1} \sum^{n}_{i=1} (z'_i - \bar{z}') (z'_i - \bar{z}')^T \text{,}
\end{equation*}
where $z'_i \coloneqq \frac{z_i}{\Vert z_i \Vert_2}$ and $\bar{z}'$ is the mean of the normalized model encoded samples.
We present the progression of eigenvalues for the image, text, and average covariance for both the sample covariance matrix \cref{fig:cov_threshold} and the kernel covariance matrix \cref{fig:cosim_cov_threshold}. We observe that this does not change the noise threshold much nor the underlying eigenspectra. 

\clearpage
\section{Random Direction Removal}
\label{apx:random_direction}

\begin{figure}[h]
    \centering
    \includegraphics[width=0.9\linewidth]{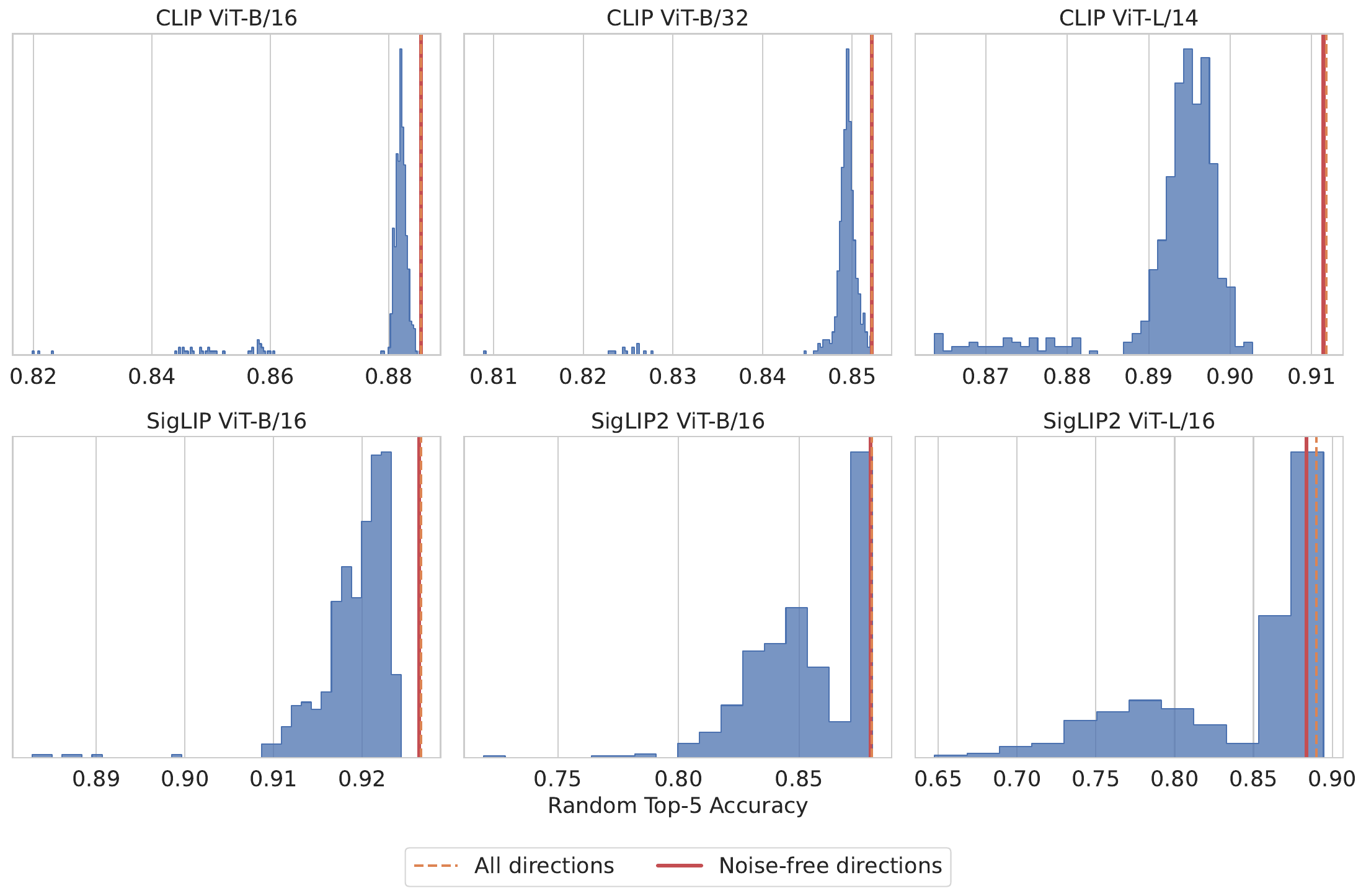}
    \caption{Distributions of Top-5 accuracy computed on VLMs zero-shot ImageNet classification. The histograms present the performance of the VLMs after removing random directions from their embeddings. The amount of removed directions is equal to the discovered noisy directions (presented in the second column of~\Cref{tab:top5_acc}).}
    \label{fig:histograms}
\end{figure}

We further extend the analysis from the main paper with a series of histograms that present how removing the random direction from VLM's embedding affects their performance. 
As indicated in~\Cref{fig:histograms}, random projection affects the performance negatively, while removing only the noise direction has no impact on the downstream task. 
An interesting exception can be observed for SigLIP2 ViT-L/16, which, in fact, can slightly benefit from random direction removal.
A contrastive behavior might be observed for CLIP ViT-L/14, which strongly suffers from removing random directions.
Still, removing specifically noise directions does not affect its performance.

\clearpage
\section{Noise Activation Examples}
\label{apx:examples}

\begin{figure}[hb]
    \centering
    \includegraphics[width=0.9\linewidth]{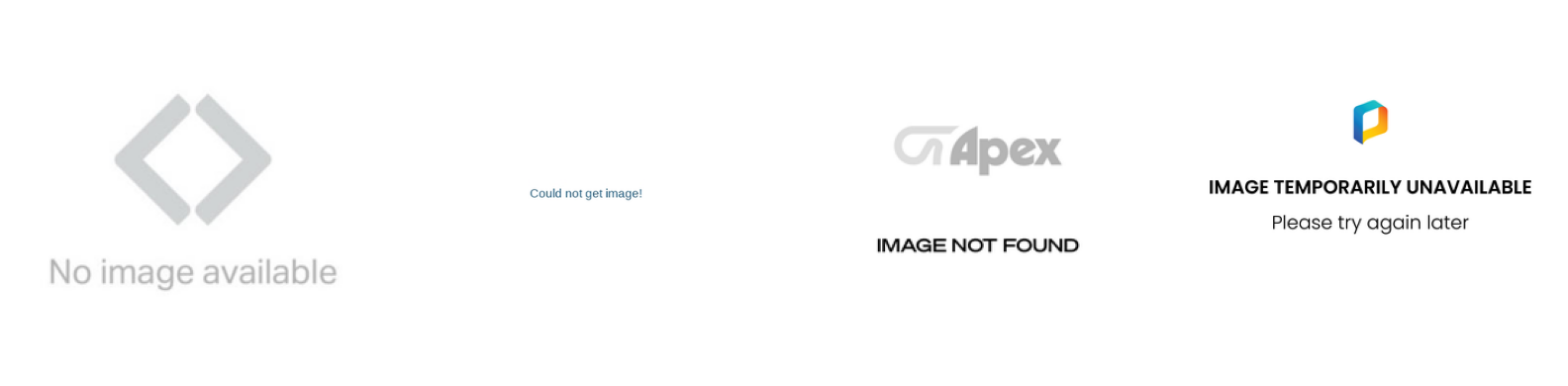}
    \caption{Examples of images most activated in noise direction.
    One of the most common images is a placeholder for unavailable images.
    The noise directions are estimated using only the LAION dataset.
    }
    \label{fig:noimage}
\end{figure}

\begin{figure}[hb]
    \centering
    \includegraphics[width=0.9\linewidth]{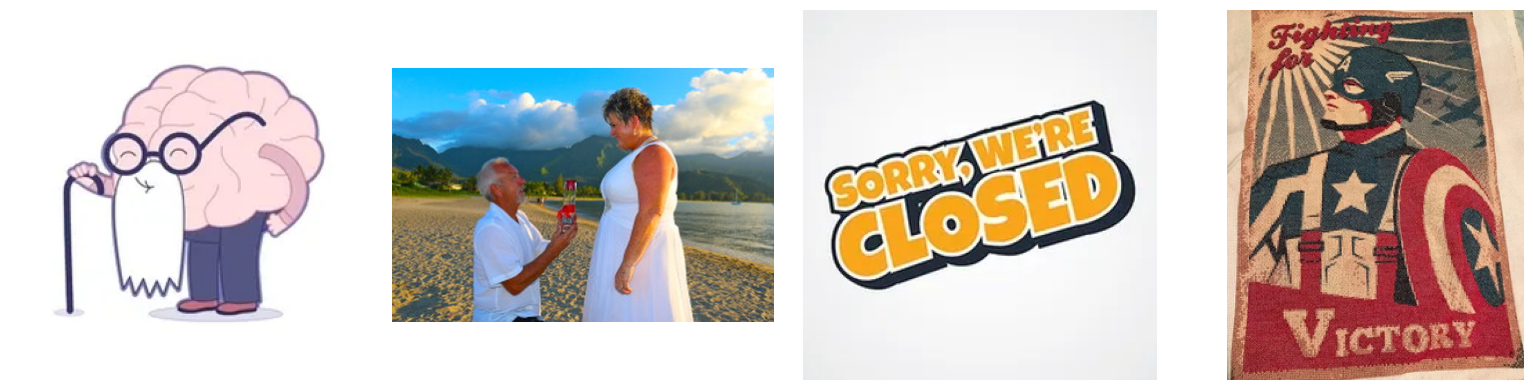}
    \caption{Examples of images most activated in noise direction.
    Images that contain content other than a placeholder for unavailable images are in the minority in this set.
    The noise directions are estimated using only the LAION dataset.}
    \label{fig:random}
\end{figure}

\begin{figure}[hb]
    \centering
    \includegraphics[width=0.9\linewidth]{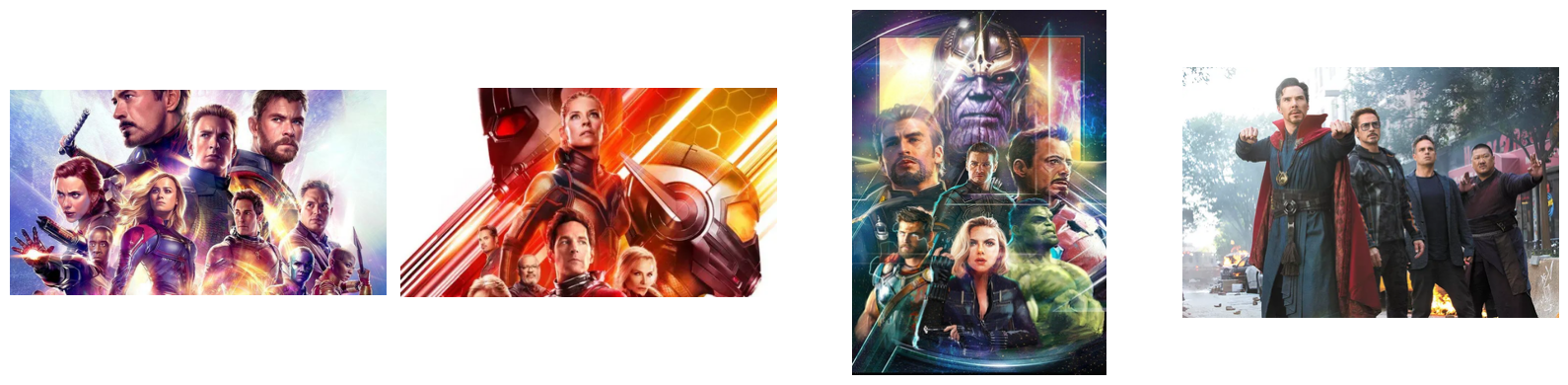}
    \caption{Examples of images most activated in noise direction.
    The noise directions are estimated using both ImageNet (images) and the LAION dataset (text).
    We argue that the topic of Marvel movies in noise space might be explained by the fact that ImageNet was collected before the latest Marvel movies (visible in the figure), which places them as out-of-distribution.
    }
    \label{fig:marvel}
\end{figure}

We present examples of samples most activated in the space spanned by the noise direction for the CLIP ViT-L/14 model.
The noise directions are found using a covariance matrix estimated on LAION's samples.
Examples of images in Figures~\ref{fig:noimage},~\ref{fig:random} are retrieved from the LAION dataset.
Most of them contain only a placeholder for unavailable images.
Notably, using covariance estimation built on images from ImageNet-1k results in multiple images from the Marvel movies, which was relased after publication of the ImageNet dataset, which is presented in Figure~\ref{fig:marvel}.
After manual investigation, we observe no Marvel-related images in the noise-activated samples when the covariance is estimated only on a LAION dataset.

\end{document}